\pdfoutput=1

\documentclass[11pt]{article}

\usepackage[preprint]{acl}

\usepackage{makecell}
\usepackage{fvextra}
\usepackage{listings}
\usepackage{times}
\usepackage{latexsym}
\usepackage{hyperref}
\usepackage{lipsum}
\usepackage{times}
\usepackage{latexsym}
\usepackage{subfig}
\usepackage{graphicx}
\usepackage{subcaption}
\usepackage{amsmath}
\usepackage{multirow}
\usepackage{tabularx}
\usepackage{enumitem}
\usepackage{pbox}
\usepackage{float}
\usepackage{hyperref}
\usepackage{setspace}
\usepackage{diagbox}
\usepackage{float}
\usepackage{subcaption}
\usepackage{caption}
\usepackage{booktabs}
\usepackage{bbding}
\usepackage{mathtools}
\usepackage{color, colortbl}
\definecolor{grey}{rgb}{0.898,0.898,0.898}
\usepackage{amssymb}%
\usepackage{makecell}
\usepackage[T1]{fontenc}

\usepackage[utf8]{inputenc}

\usepackage{microtype}

\usepackage{inconsolata}

\usepackage{graphicx}
\usepackage{xcolor}
\usepackage{amsmath}
\usepackage{cleveref}
\usepackage{tabularx}
\usepackage{lipsum} 
\usepackage{ragged2e}
\usepackage{pifont}
\usepackage{tikz}
\usetikzlibrary{arrows.meta, positioning, shapes.misc, fit, calc}

\newcommand{\cmark}{\ding{51}} 
\newcommand{\xmark}{\ding{55}} 
\newcolumntype{P}[1]{>{\RaggedRight\hspace{0pt}}p{#1}}

\definecolor{headerblue}{RGB}{200, 220, 255}
\definecolor{rowlightblue}{RGB}{240, 245, 255}
\definecolor{rowlightgray}{RGB}{245, 245, 245}
\definecolor{typegreen}{RGB}{180, 210, 180}

\usepackage{tcolorbox}

\title{Disambiguation in Conversational Question Answering in the Era of LLMs and Agents: A Survey}

\author{Md Mehrab Tanjim$^1$, Yeonjun In$^2$\thanks{Joint First Author}, Xiang Chen$^1$, Victor S. Bursztyn$^1$, \\
\textbf{Ryan A. Rossi$^1$, Sungchul Kim$^1$, Guang-Jie Ren$^3$, Vaishnavi Muppala$^3$,}\\
\textbf{Shun Jiang$^3$, Yongsung Kim$^3$, Chanyoung Park$^2$}\\
$^1$Adobe Research, $^2$KAIST, $^3$Adobe Inc.\\
  \texttt{\{tanjim, xiangche, soaresbu, ryrossi, sukim, }\\
  \texttt{gren, mvaishna, shunj, yongsungk\}@adobe.com} \\
  \texttt{\{yeonjun.in, cy.park\}@kaist.ac.kr} \\}

\begin{document}
\maketitle
\begin{abstract}

Ambiguity remains a fundamental challenge in Natural Language Processing (NLP) due to the inherent complexity and flexibility of human language. With the advent of Large Language Models (LLMs), addressing ambiguity has become even more critical due to their expanded capabilities and applications. In the context of Conversational Question Answering (CQA), this paper explores the definition, forms, and implications of ambiguity for language driven systems, particularly in the context of LLMs. We define key terms and concepts, categorize various disambiguation approaches enabled by LLMs, and provide a comparative analysis of their advantages and disadvantages. We also explore publicly available datasets for benchmarking ambiguity detection and resolution techniques and highlight their relevance for ongoing research. Finally, we identify open problems and future research directions, especially in agentic settings, proposing areas for further investigation. By offering a comprehensive review of current research on ambiguities and disambiguation with LLMs, we aim to contribute to the development of more robust and reliable LLM-based systems.

\end{abstract}

\section{Introduction}
The inherent ambiguity in natural language communication presents a fundamental challenge in human-AI interactions, especially in conversational systems. 
Modern AI Assistants, such as Adobe's AEP AI Assistant\footnote{\href{https://business.adobe.com/products/sensei/ai-assistant.html}{business.adobe.com/products/sensei/ai-assistant.html}} and Amazon's Rufus\footnote{\href{https://www.aboutamazon.com/news/retail/how-to-use-amazon-rufus}{aboutamazon.com/news/retail/how-to-use-amazon-rufus}}, must navigate these ambiguities through advanced language understanding mechanisms. The ability to accurately determine the intended meaning of a term or phrase within a given context is fundamental to enhancing the performance of such conversational systems. This mirrors human cognitive behavior, where communicators must anticipate potential misunderstandings, while recipients engage in active disambiguation through contextual analysis \cite{anand2023context}, clarifying questions \cite{zamani2020generating, zhang2024clamber}, and continuous interpretation refinement \cite{zukerman2002lexical, jones2006generating}. 

\begin{figure}[tbp]
    \centering
    \includegraphics[width=0.85\linewidth]{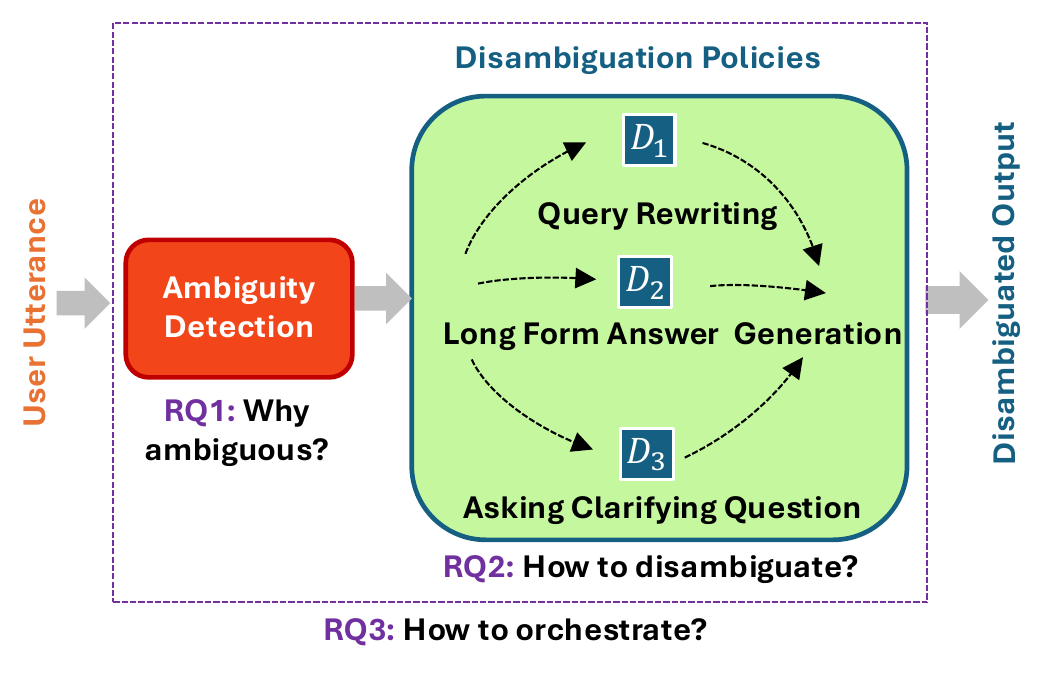} 
    \vspace{-0.5cm}
    \caption{Broadly, we categorize the existing literature to answer three major research questions (RQs), namely, why ambiguous (RQ1), how to disambiguate (RQ2), and how to orchestrate (RQ3).}
    \label{fig:teaser}
    \vspace{-0.7cm}
\end{figure}
The advent of Large Language Models (LLMs) has further underscored the importance of understanding and resolving ambiguity to enhance the performance and reliability of language understanding systems. As LLMs become increasingly integral to applications, such as search engines or Information Retrieval (IR) \cite{anand2023context, ma2023query}, Conversational Question Answering (CQA) \cite{zhang2020dialogpt, thoppilan2022lamda, xu2023baize}, automated text summarization \cite{kurisinkel2023llm, zakkas2024sumblogger} and so on, their ability to manage ambiguous language is essential for effective communication and user satisfaction. This is because their utility can often be compromised by ambiguous user queries, which can lead to incorrect or irrelevant outputs \cite{kuhn2022clam, deng2023rethinking}.

While disambiguation techniques have witnessed significant advancements over recent decades, driven by sophisticated algorithms \cite{raganato-etal-2017-neural, zhang2018towards, rao2018learning, rao2019answer, xu-etal-2019-asking, aliannejadi2019asking, kumar2020clarq, min2020ambigqa, zamani2020generating, guo2021abgcoqa, kuhn2022clam, lee2023asking}, the inherent complexity of natural language and the need for large annotated corpora has been continuing to pose substantial challenges. For these reasons, an emerging and active area of research is to explore the capacity of LLMs themselves to identify and resolve ambiguous queries  \cite{liu2023we, mehrparvar2024detecting, zhang2023clarify, zhang2024clamber, anand2023context}. 
While LLM-based disambiguation techniques are gaining popularity, the field lacks a systematic analysis and categorization of existing methods. This paper addresses that gap by surveying current LLM-based approaches for ambiguity detection and disambiguation, outlining their underlying principles, strengths, and limitations. 
{Among the NLP tasks, we primarily focus on CQA as this task seems to be prominent in majority of the use-cases.}

\medskip\noindent\textbf{Organization of this Survey.}
We structure this survey around three core research questions (see \Cref{fig:teaser}):
RQ1: Why do ambiguities arise in language, and how can we detect them?
RQ2: How can we disambiguate, particularly using LLMs?
RQ3: How can we automate disambiguation strategies in real-world applications?
Section~\ref{sec:ambiguity} addresses RQ1 by defining key concepts, presenting a taxonomy, and reviewing ambiguity detection methods. Section~\ref{sec:disambiguation} tackles RQ2 by categorizing LLM-based disambiguation approaches and analyzing their strengths and weaknesses. To support these, Section~\ref{sec:benchmarks} surveys relevant public datasets used for benchmarking. Finally, Section~\ref{sec:open-problems-challenges} explores open challenges and outlines future directions, centering on RQ3: how to orchestrate disambiguation effectively in practice.

\section{\textit{Why Ambiguous?}}\label{sec:ambiguity}
\subsection{Definition of Ambiguity}

Ambiguous queries are typically those that have multiple distinct meanings, insufficiently defined subtopics \cite{clarke2009overview}, syntactic ambiguities \cite{schlangen2004causes}, for which a system struggles to interpret accurately, resulting in inappropriate or unclear answers \cite{keyvan2022approach}. 
These ambiguities can arise at lexical, syntactic, or semantic levels, motivating the development of various taxonomies, which we present in the next section.

\begin{table*}[ht]
\centering
\resizebox{\textwidth}{!}{
\begin{tabular}{P{2cm}P{3.5cm}P{8cm}P{7.5cm}}
\toprule
\rowcolor{headerblue}
\textbf{Literature} & \multicolumn{3}{c}{\textbf{Taxonomy}} \\
\cmidrule(lr){2-4}
\rowcolor{headerblue}
& \textbf{Type} & \textbf{Definition Provided by the Literature} & \textbf{Example Given} \\ 
\midrule

\rowcolor{rowlightblue}
\cellcolor{white}\citet{tanjim2025detecting} & 
\cellcolor{typegreen}Pragmatic & The meaning of a sentence depends on the context, reference, or scope. & \textit{``How \underline{many} do I have?''} \\
\rowcolor{rowlightblue}
& \cellcolor{typegreen}Syntactic & The structure of a sentence is incomplete or allows for multiple interpretations. & \textit{``\underline{Business event}''} \\
\rowcolor{rowlightblue}
& \cellcolor{typegreen}Lexical & The meaning of the word/term is not clear or has multiple interpretations. & \textit{``Are we removing \underline{abc123} from \underline{XYZ}?''} \\

\midrule

\rowcolor{rowlightgray}
\cellcolor{white}\citet{zhang2024clamber} & 
\cellcolor{typegreen}Unfamiliar & Query contains unfamiliar entities or facts. & \textit{``Find the price of \underline{Samsung Chromecast}.''} \\
\rowcolor{rowlightgray}
& \cellcolor{typegreen}Contradiction & Query contains self contradictions. & \textit{``Output `X' if the sentence contains [category withhold] and `Y' otherwise. The critic is in the restaurant.>X. The butterfly is in the river.>Y. \underline{The boar is in the theatre}?''} \\
\rowcolor{rowlightgray}
& \cellcolor{typegreen}Lexical & Query contains terms with multiple meanings. & \textit{``Tell me about the \underline{source of Nile}.''} \\
\rowcolor{rowlightgray}
& \cellcolor{typegreen}Semantic & Query lacks context leading to multiple interpretations. & {``When did \underline{he} land on the moon?''} \\
\rowcolor{rowlightgray}
& \cellcolor{typegreen}Aleatoric & Query output contains confusion due to missing personal/temporal/spatial/task-specific elements. & \textit{``How many goals did Argentina score in \underline{the World Cup}?''} \\

\midrule

\rowcolor{rowlightblue}
\cellcolor{white}\citet{liu2023we} & 
\cellcolor{typegreen}Pragmatic & Literal and pragmatic interpretations are present. & \textit{``\underline{I’m afraid} the cat was hit by a car.''} \\
\rowcolor{rowlightblue}
& \cellcolor{typegreen}Lexical & A lexical item has different senses. & \textit{``John and Anna are \underline{married}.''} \\
\rowcolor{rowlightblue}
& \cellcolor{typegreen}Syntactic & Different syntactic parses lead to different interpretations. & \textit{``This seminar is full now, but \underline{interesting seminars} are being offered next quarter too.''} \\
\rowcolor{rowlightblue}
& \cellcolor{typegreen}Scopal & Ambiguity from the relative scopal order of quantifiers or the scope of particular modifiers. & \textit{``The novel has been banned \underline{in many schools} because of its explicit language.''} \\
\rowcolor{rowlightblue}
& \cellcolor{typegreen}Coreference & Ambiguous coreference. & \textit{``It is currently March, and they plan to schedule their wedding for \underline{next December}.''} \\

\midrule

\rowcolor{rowlightgray}
\cellcolor{white}\citet{zhang2023clarify} & 
\cellcolor{typegreen}Word-Sense Disambiguation & Word-sense disambiguation for named entities, also commonly surfaces as entity linking ambiguities. & \textit{``Who wins at the end of \underline{friday night lights}?''} \\
\rowcolor{rowlightgray}
& \cellcolor{typegreen}Literal vs.~Implied Interpretation & A question literally means something different from what the user probably meant to ask. & \textit{``The cake was so dry, it was like \underline{eating sand}.''} \\
\rowcolor{rowlightgray}
& \cellcolor{typegreen}Multiple Valid Outputs & Ambiguity due to multiple valid outputs. & \textit{``\underline{When} did west germany win the world cup?''} \\

\bottomrule
\end{tabular}
}
\caption{Here, we present several taxonomies exactly as they appear in the existing literature, along with their definitions and examples (ambiguous parts of the text are underlined). As can be seen there are redundancies in these definitions, highlighting the need for a unified taxonomy.}
\label{tab:taxonomy}
\vspace{-0.2cm}
\end{table*}

\subsection{Taxonomy of Ambiguity}
Existing literature approaches the taxonomy of ambiguities in various ways, often influenced by specific use-cases, public datasets, or the scope defined for new data collection. For instance, \citet{tanjim2025detecting} focuses on industrial conversation question answering, while \citet{zhang2024clamber} examine ambiguities through public datasets. Additionally, \citet{liu2023we} define their own criteria for collecting new datasets, further diversifying the landscape of ambiguity taxonomies. This complexity is compounded by the various NLP tasks to which these taxonomies are applied. For example, Natural Language Inference (NLI), Question Answering (QA), and Machine Translation (MT) each have unique requirements and interpretations of ambiguity, as explored by \citet{zhang2023clarify}. Consequently, different taxonomies have emerged from these diverse focuses. Moreover, the same example can be treated differently across various studies. For instance, \citet{zhang2024clamber} categorized the example \textit{``Real name of \underline{gwen stacy} in amazing spiderman?"} as an Aleatoric `What' type of ambiguity. In contrast, \citet{zhang2023clarify} classified this as a `Literal vs.~Implied interpretation' ambiguity. This discrepancy underscores the need for a unified approach to taxonomy.

In \Cref{tab:taxonomy}, we present a comparative analysis of these taxonomies to highlight common grounds despite their differences. To cater to broader applications and provide clarity, we propose simplifying existing taxonomies into three overarching categories. We argue that these categories can encompass all existing taxonomies, irrespective of the underlying tasks, thereby offering a more cohesive framework for understanding ambiguities.

\noindent\textbf{Syntactic Ambiguity:} When a sentence can be parsed in different ways \citep{church1982coping, wasow2015ambiguity}. For example, \textit{`I saw the man with a telescope.'} Here the ambiguity arises because it could be interpreted in two ways: did the speaker see the man `with the telescope' or did the speaker see `the man' using the telescope? This taxonomy is listed in both \citet{tanjim2025detecting} and \citet{liu2023we}, but it seems to be missing in the other two.

\noindent \textbf{Semantic Ambiguity:} When a sentence is grammatically correct but semantically unclear, due to ambiguity in a word, phrase, or the overall interpretation. The more common case involves ambiguity at the word or phrase level, 
often referred to as \textit{lexical ambiguity} \citep{navigli2009word, beekhuizen2021probing},
where a term has multiple possible meanings. 
As shown in \Cref{tab:taxonomy}, 
this type is listed
across most prior work, with the exception of \citet{zhang2023clarify}, where they mention it as `word sense disambiguation.' Similarly, the `Unfamiliar' category in \citet{zhang2024clamber} aligns with this type, as unknown words are inherently open to interpretation until contextual or domain-specific knowledge is applied. Beyond word-level issues, semantic ambiguity can also stem from interpretive variation at the sentence level. This includes the usage of literal vs.~pragmatic words as mentioned by \citet{liu2023we}, who refer to it as pragmatic ambiguity, and `Literal vs.~Implied Interpretations' by \citet{zhang2023clarify}. The `Figurative' type in \citet{liu2023we} also falls into this category, as does the `Contradiction' category in \citet{zhang2024clamber} because of conflicts with the semantics of previous statements.

\noindent\textbf{Contextual Ambiguity:} When the context of the conversation is missing or the answers could be multiple unless no specific context is given (e.g., what/when/where/who type of questions without context) \citep{sperber1986relevance, huang2017pragmatics}. \citet{tanjim2025detecting} name this as pragmatic ambiguity, whereas it is listed as `Semantics' in \citet{zhang2024clamber} and as `Aleatoric', `Coreference' and `Scopal' in \citet{liu2023we}, and as `Multiple Valid Outputs' in \citet{zhang2023clarify}. Meanwhile, `Knowledge Conflict`, as described by \citet{neeman2022disentqa,shaier2024adaptive}, also aligns with this type, occurring when a question lacks specific context, such as temporal or locational cues, causing retrieval-augmented models to face conflicts between retrieved and parametric knowledge.

\begin{table}[t]
\centering
\resizebox{\columnwidth}{!}{
\begin{tabular}{p{2cm}|p{2.8cm}|p{2.4cm}|p{3.4cm}}
\toprule
\rowcolor{headerblue}
\textbf{Literature} & \textbf{Approach} & \textbf{Inputs} & \textbf{Ambiguity Type} \\
\midrule

\rowcolor{gray!10}
\citet{trienes2019identifying} & Logistic regression + features & Q, tags, similar Qs & Syntactical \\

\rowcolor{gray!10}
\citet{dhole2020resolving} & BiLSTM classifier & Dialogue, intents & Contextual \\

\rowcolor{cyan!10}
\citet{guo2021abgcoqa} & BERT classifier & Conv., passage & Semantic, Contextual \\

\rowcolor{cyan!10}
\citet{lee2023asking} & BERT classifier & Q, passages & Contextual \\

\rowcolor{cyan!10}
\citet{tanjim2025detecting} & ST + rules + features & Q only & Syntactical, Semantic, Contextual \\

\rowcolor{pink!10}
\citet{kuhn2022clam} & Prompted LLM & Q Only & Contextual \\

\rowcolor{pink!10}
\citet{zhang2024clamber} & Prompted LLM & Q, context (optional) & Semantic, Contextual \\

\rowcolor{pink!10}
\citet{zhang2023clarify} & LLM + CoT by ambiguity type & Q, prompt schema & Semantic, Contextual \\

\rowcolor{pink!10}
\citet{kim2024aligning} & LLM + uncertainty signals & Q only & Semantic, Contextual \\

\bottomrule
\end{tabular}
}
\caption{Summary of ambiguity detection methods. Shaded by method type: traditional (gray), LM (cyan), LLM (pink). Here, ST= Sentence Transformer, Q=Question, Conv.= Conversation.}
\label{tab:amb_det_small}
\vspace{-0.3cm}
\end{table}

\subsection{Ambiguity Detection}

The body of work for detecting ambiguity can be broadly categorized into three major groups: traditional methods (not language model-based), language model-based methods, and large language model (LLM)-based methods. In \Cref{tab:amb_det_small}, we summarize each method’s approach, model inputs, and the types of ambiguity it addresses based on our taxonomy. We give more details below.

\medskip\noindent\textbf{Traditional Methods:}
Early research into ambiguity detection primarily concentrated on binary classification methodologies. A significant contribution in this domain was made by \citet{trienes2019identifying}, who used logistic regression on features from similar questions in community QA forums. Their model and features targeted queries that have a defect in their structure, thereby focusing on \textit{syntactical ambiguity}. While offering interpretability, their scope was limited to single-turn QA and did not account for other ambiguity types such as semantic or contextual ambiguities in dialogue-based settings. To address some of these limitations, \citet{dhole2020resolving} proposed a two-stage approach for resolving ambiguous user intents in task-oriented dialogue. Their work falls under \textit{contextual ambiguity}, as their classifier disambiguates underspecified user intents. 

\medskip\noindent\textbf{Language Model-Based Methods:}
In the realm of language model-based methods, \citet{guo2021abgcoqa} introduced Abg-CoQA, a benchmark dataset and framework for ambiguity detection and clarifying question generation in conversational QA. Their model addressed both \textit{semantic} and \textit{contextual} ambiguities owing to their framing ambiguity detection as a QA classification task (thus capable of understanding the semantic ambiguity). However, even with BERT-based models, performance remained low (23.6\% F1). Similarly, \citet{lee2023asking} proposed a BERT-based classifier to detect ambiguity given a passage, but their model also exhibited low performance. Their work primarily focused on \textit{contextual ambiguity}, where a question can lead to multiple valid answers without further specification. A more recent study by \citet{tanjim2025detecting} employed Sentence Transformers with handcrafted rules and features to detect all three ambiguity types—\textit{syntactic}, \textit{semantic}, and \textit{contextual}—demonstrating that explicit modeling of ambiguity categories can improve detection.

\medskip\noindent\textbf{LLM-Based Methods:}
With the advent of large language models (LLMs), ambiguity detection has increasingly shifted toward prompt-based methods. \citet{kuhn2022clam} demonstrated that LLMs could be prompted to decide whether to answer a query or ask for clarification. Their method targeted primarily \textit{contextual ambiguity}, especially in cases of underspecified user queries. \citet{zhang2024clamber} introduced CLAMBER, a benchmark with a taxonomy of eight ambiguity types. They showed that LLMs can identify certain \textit{semantic} (e.g., lexical or referential ambiguity) and \textit{contextual} ambiguities, but struggle with systematic disambiguation. \citet{zhang2023clarify} proposed a prompting method that asks the model to reason about ambiguity types before generating a clarifying question. Their framework covers both \textit{semantic} and \textit{contextual ambiguity}, aligning clarification strategies with the predicted ambiguity type. Finally, \citet{kim2024aligning} presented a method where LLMs use their internal uncertainty to decide whether a query is ambiguous. Their alignment framework quantifies information gain through clarification, capturing \textit{semantic} ambiguities (e.g., polysemous terms) and \textit{contextual} ones (e.g., missing scope or domain).

Despite the flexibility of LLMs, these works collectively show that ambiguity detection—particularly fine-grained distinctions among types—remains a complex problem. We will revisit these challenges in \Cref{sec:open-problems-challenges}.

\begin{figure*}[tbp]
    \centering
    \includegraphics[width=0.9\linewidth]{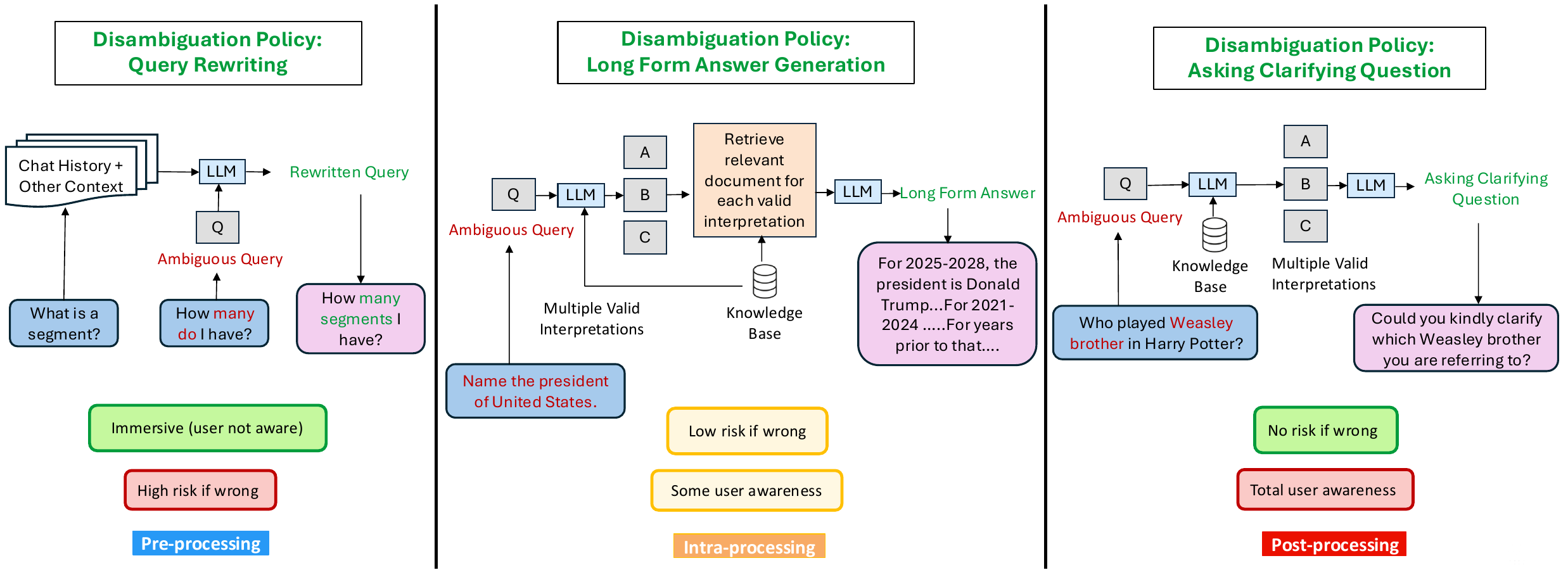} 
    \caption{We find existing disambiguation efforts using LLMs broadly fall into these three major categories: Left. Query Rewriting, Middle. Long Form Answer Generation, Right. Asking Clarifying Questions. These policies have different requirements and also work in different processing steps in CQA pipeline, resulting in unique advantages and disadvantages for each approach. We highlight a couple here and provide a more comprehensive list in \Cref{tab:comparison}.}
    \label{fig:dismabiguation}
\end{figure*}
\section{\textit{How To Disambiguate?}}\label{sec:disambiguation}
In the era of LLMs, disambiguation is gaining increasing attention due to their extensive world knowledge and advanced capabilities, surpassing traditional and smaller language models. However, current research in this area often lacks systematic categorization and tends to address various aspects in isolation. To that end, in this paper, we argue existing disambiguation works fall in three major policies, which we present in \Cref{fig:dismabiguation}. We describe each of them below.

\subsection{Query Rewriting (QR)}
Query rewriting (QR) represents a wide span of techniques that transforms ambiguous or unclear user queries into well-defined, comprehensive expressions \cite{carpineto2012survey}. Early work focused on query expansion \cite{carpineto2012survey, lavrenko2017relevance}, contextual rephrasing \cite{zukerman2002lexical}, and synonym-based augmentation \cite{jones2006generating}. Prior to LLM, research demonstrates significant advances in neural query rewriting through supervised learning approaches \cite{elgohary2019can, anantha2021open} and reinforcement learning frameworks \cite{vakulenko2021question}. Other innovations have explored explicit reasoning patterns \cite{qian2022explicit} achieving good performance in transforming ambiguous queries into precise, answerable questions. 

The emergence of LLMs has enabled more advanced query reformulation, moving beyond term-based edits to deeper semantic understanding and contextual refinement across downstream tasks \cite{wang2023query2doc}. Recent research works, such as \citet{ma2023query,jagerman2023query}, have demonstrated the efficacy of LLM-based query reformulation in zero-shot and few-shot settings, particularly valuable when domain-specific training data is scarce. The principled way of QR is shown in \Cref{fig:dismabiguation} (Left), where an LLM is prompted with previous chat history and other relevant information as context. Some advanced prompting, such as \citet{ye2023enhancing} also includes ``rewrite-then-edit'' framework. Apart from prompting, LLMs also have been fined-tuned \cite{peng2024large} or used to generate Supervised Fine Tuning (SFT) dataset to improve QR model either through a re-ranker \cite{mao2024rafe} or preference optimization \cite{zhang2024adaptive}. 

\subsection{Long Form Answer Generation (LFAG)}
Generating long-form answers to ambiguous questions involves presenting all valid interpretations alongside their corresponding answers. For instance, the question ``Who has the highest goals in world football?'' can refer to either men’s or women’s football. A well-structured response would be: ``Ali Daei holds the record in men’s football, while Christine Sinclair does in women’s football.'' As shown in \Cref{fig:dismabiguation} (Middle), this task typically comprises three steps: \textbf{1)} Disambiguating the question, \textbf{2)} Answering each interpretation, and \textbf{3)} Consolidating the results into a single, coherent response. Early methods streamlined these steps into a single model inference. \citet{stelmakh-etal-2022-asqa} finetuned T5 to directly produce long-form answers. More recent LLM-based approaches, such as \citet{gao2023enabling}, show that few-shot prompting can be similarly effective without fine-tuning. To reduce reasoning load, \citet{amplayo2022query} proposed a two-step method: first inferring multiple interpretations, then generating a long-form answer from them. RAC \cite{kim2023tree} introduced retrieval-augmented disambiguation to generate answers with supporting evidence (Steps 1--2), while ToC \cite{kim2023tree} extended this via iterative retrieval to capture overlooked interpretations, trading off efficiency. DIVA \cite{in2024diversify} improved efficiency by modeling a reasoning chain that compresses this process into a single step, maintaining performance while reducing complexity.

\begin{table}[t]
\centering
\resizebox{\columnwidth}{!}{
\begin{tabular}{lccc}
\toprule
\textbf{Technique} & \textbf{Syntactic} & \textbf{Semantic} & \textbf{Contextual} \\
\midrule
QR & \cmark & \cmark & \cmark \\
LFAG & \xmark & \cmark & \cmark \\
ACQ & \xmark & \cmark & \cmark \\
\bottomrule
\end{tabular}
}
\caption{Disambiguation techniques and the types of ambiguity they are equipped to handle.}
\label{tab:technique-vs-ambiguity}
\vspace{-0.4cm}
\end{table}

\begin{table*}[ht]
\centering
\resizebox{\textwidth}{!}{
\begin{tabular}{
>{\arraybackslash}p{2cm}|
>{\arraybackslash}p{2cm}|
>{\arraybackslash}p{1.8cm}|
>{\arraybackslash}p{3cm}|
>{\arraybackslash}p{2.1cm}|
>{\arraybackslash}p{1.9cm}|
>{\arraybackslash}p{5.5cm}}
\toprule
\rowcolor{headerblue}
\textbf{Paper} & \textbf{Name} & \textbf{Domain} & \textbf{Core Unit} & \textbf{Scale} & \textbf{\# Ambiguous} & \textbf{Link} \\

\midrule
\rowcolor{gray!20}\multicolumn{7}{c}{\textbf{\textit{Ambiguity Detection and Asking Clarifying Question}}} \\

\rowcolor{yellow!15}
\citet{xu-etal-2019-asking} & \textbf{CLAQUA} & Open-domain & Q w/ Ans. (ST + MT) & 17K + 22K & 7K + 9K & \href{https://github.com/msra-nlc/MSParS_V2.0}{github.com/msra-nlc/MSParS\_V2.0} \\

\rowcolor{pink!15}
\citet{kumar2020clarq} & \textbf{ClarQ} & Stack Exchange & Q w/ Context & 6M & 2M & \href{https://github.com/vaibhav4595/ClarQ}{github.com/vaibhav4595/ClarQ} \\

\rowcolor{cyan!10}
\citet{min2020ambigqa} & \textbf{AmbigNQ} & Wikipedia & Q w/ Ans. (Tr/Vl/Te) & 10K / 2K / 2K & 4K / 1K / 1K & \href{https://nlp.cs.washington.edu/ambigqa/}{nlp.cs.washington.edu/ambigqa} \\

\rowcolor{yellow!15}
\citet{guo2021abgcoqa} & \textbf{Abg-CoQA} & Stack Exchange & P + Q & 4K + 8K & 800+ / 900+ & \href{https://github.com/MeiqiGuo/AKBC2021-Abg-CoQA/tree/main/abg-coqa}{github.com/MeiqiGuo/AKBC2021-Abg-CoQA} \\

\rowcolor{pink!15}
\citet{aliannejadi-etal-2021-building} & \textbf{ClariQ} & TREC, Qulac & Conv. + Clar.Q & 11K + 1M & Rated & \href{https://github.com/aliannejadi/ClariQ}{github.com/aliannejadi/ClariQ} \\

\rowcolor{cyan!10}
\citet{deng2022pacific} & \textbf{PACIFIC} & TAT-QA & Conv. + Q w/ Context \& Ans. & 2K + 19K & 2K & \href{https://github.com/dengyang17/PACIFIC}{github.com/dengyang17/PACIFIC} \\

\rowcolor{cyan!10}
\citet{lee2023asking} & \textbf{CAmbigNQ} & AmbigNQ & Clar.Q + Ans. + P & 4K + 400+ + 400+ & All Ambig. & \href{https://github.com/DongryeolLee96/AskCQ}{github.com/DongryeolLee96/AskCQ} \\

\rowcolor{cyan!10}
\citet{zhang2024clamber} & \textbf{CLAMBER} & Mixed & Q w/ Context & 12K & 5K & \href{https://github.com/zt991211/CLAMBER}{github.com/zt991211/CLAMBER} \\

\midrule
\rowcolor{gray!20}\multicolumn{7}{c}{\textbf{\textit{Query Rewriting}}} \\

\rowcolor{cyan!10}
\citet{elgohary2019can} & \textbf{CANARD} & QUAC & Q + Rewrite & 40K + 40K & N/A & \href{https://sites.google.com/view/qanta/projects/canard}{canard.qanta.org} \\

\rowcolor{pink!15}
\citet{anantha2021open} & \textbf{QReCC} & QUAC, NQ, TREC-C & Conv. + Q + Rewrite & 13K + 80K + 80K & N/A & \href{https://github.com/apple/ml-qrecc}{github.com/apple/ml-qrecc} \\

\midrule
\rowcolor{gray!20}\multicolumn{7}{c}{\textbf{\textit{Long Form Answer Generation}}} \\

\rowcolor{cyan!10}
\citet{stelmakh-etal-2022-asqa} & \textbf{ASQA} & Wikipedia, AmbigNQ & Q w/ LF Ans. (Tr/Vl/Te) & 4K / 900+ / 1K & All Ambig. & \href{https://github.com/google-research/language/tree/master/language/asqa}{github.com/google-research/language} \\
\bottomrule
\end{tabular}}
\caption{Publicly available datasets for benchmarking ambiguity in QA, covering both ambiguous and non-ambiguous cases (except ASQA, CANARD, QReCC). Rows are task-grouped and color-coded by size: large (\textcolor{pink!60}{pink}), medium (\textcolor{cyan!60}{cyan}), small (\textcolor{yellow!50!orange}{yellow}). "Core Unit" abbreviates data structure: Tr=Train, Vl=Val, Te=Test, P=Passage, Q=Question, Ans.=Answer, Clar.Q=Clarifying Q., Conv.=Conversation, LF=Long Form, Context=Passage/Table/Post (depends on the dataset), Rated=All questions rated from 1 (clear) to 4 (ambiguous).}
\label{tab:ambig-datasets-cleaned}
\vspace{-0.1cm}
\end{table*}

\subsection{Asking Clarifying Question (ACQ)} 
This is one of the most extensively studied disambiguation policies, with approaches ranging from rule-based prompts (e.g., ``Did you mean A or B?'' \cite{coden2015did}, ``What do you want to know about QUERY?'' \cite{zamani2020generating}, or category-based options \cite{lee2023asking}) to traditional machine learning \cite{zhang2018towards, rao2018learning, rao2019answer} and language model-based methods \cite{xu-etal-2019-asking, aliannejadi2019asking}. Several works also introduce new datasets \cite{xu-etal-2019-asking, kumar2020clarq, min2020ambigqa, guo2021abgcoqa}, discussed further in \Cref{sec:benchmarks}. However, these methods often struggle with complex queries and rely on annotated corpora, which could be difficult to obtain.

With the advent of LLMs, recent studies have leveraged prompt-based approaches \cite{kuhn2022clam, deng-etal-2023-prompting, zhang2024clamber}, typically employing zero-shot or few-shot Chain-of-Thought (CoT) prompting strategies. These methods mirror the Long-form Answer Generation pipeline but focus on analyzing multiple valid interpretations to generate clarifying questions, as shown in \Cref{fig:dismabiguation} (Right). Like QR, they reduce the need for domain-specific data and can be training-free while supporting complex question structures. Some works adopt a two-stage pipeline: first detecting ambiguity, then generating suitable clarification questions. For instance, \citet{zhang2023clarify} proposed an innovative uncertainty estimation technique for ambiguity detection that quantifies intent entropy through simulated user-assistant interactions. Finally, similar to QR, LLMs can be also be fine-tuned to generate clarifying questions. For example, \citet{zhang2024modeling, kim2024aligning} fine-tuned various LLMs, such as \texttt{Llama-2-7B} \cite{touvron2023llama}, \texttt{Gemma-7B} \cite{team2024gemma}, and \texttt{Llama-3-8B} \cite{dubey2024llama}.

\Cref{tab:technique-vs-ambiguity} summarizes how disambiguation techniques address different ambiguity types. \textbf{QR} handles all three by reformulating queries to fix syntactic issues, resolve semantic confusion through inferred interpretations, and incorporate missing contextual details from prior conversation. \textbf{LFAG} handles semantic and contextual ambiguity by presenting multiple plausible interpretations, including those that differ semantically as well as those that are plausible when considering different contexts. 
\textbf{ACQ} resolves semantic and contextual ambiguity by explicitly asking the user to confirm among similar options or supply missing information. While QR might look most appealing for its broad coverage, it still faces key challenges such as semantic drift \cite{anand2023context} and practical concerns like latency, cost, and error propagation in production \cite{tanjim2025detecting}. We will discuss the strengths and limitations of each approach further in \Cref{sec:open-problems-challenges}.

\section{Benchmarks}\label{sec:benchmarks}
To evaluate disambiguation strategies, prior work has introduced task-specific benchmark datasets and metrics, which we describe below.

\noindent{\textit{\textbf{Ambiguity Detection and ACQ.}}} Most existing datasets related to ambiguity fall into the category of detecting the need for clarification and necessary disambiguation by asking clarification questions. Notable datasets in this area include CLAQUA \cite{xu-etal-2019-asking}, ClarQ \cite{kumar2020clarq}, AmbigNQ \cite{min2020ambigqa}, ClariQ \cite{aliannejadi2020convai3}, Abg-CoQA \cite{guo2021abgcoqa}, PACIFIC \cite{deng2022pacific}, CAmbigNQ \cite{lee2023asking}, and CLAMBER \cite{zhang2024clamber}. These corpora exhibit significant variation across several dimensions, each contributing uniquely to the understanding of ambiguities in dialogue systems, as listed in \Cref{tab:ambig-datasets-cleaned}. Among them, the CLAMBER benchmark \cite{zhang2024clamber} has emerged as the first comprehensive evaluation benchmark for LLM-based ambiguity detection and ACQ, providing valuable insights into the current limitations of LLM-based approaches and establishing baseline metrics for future research. Statistics for all these datasets, along with their corresponding URLs, appear in \Cref{tab:ambig-datasets-cleaned}. Metrics typically used for ambiguity detection include classification metrics such as Precision, Recall, F1, Accuracy, and AUROC score \cite{zhang2024clamber, tanjim2025detecting}. For ACQ, the metrics are usually automatic text evaluation metrics, such as BLEU \cite{papineni2002bleu} or ROUGE \cite{lin2004rouge}. However, some studies criticize the limitations of these metrics and favor human judgment instead \cite{zamani2020generating}.

\noindent{\textit{\textbf{Query Rewriting.}}} There are two prominent benchmark datasets for evaluating the quality of rewritten queries. The pioneering dataset in this area is CANARD \cite{elgohary2019can}, which includes questions with context and their rewritten versions. This was followed by QReCC \cite{anantha2021open}, where each user question is accompanied by a human-rewritten query, and answers to questions within the same conversation may be distributed across multiple web pages. Notably, QReCC is used in recent LLM-based QR approaches such as \citet{ye2023enhancing} and \citet{zhang2024adaptive}. Both of these datasets, along with their statistics and URLs, are listed in \Cref{tab:ambig-datasets-cleaned}. It is important to note that, unlike datasets related to ACQ, these datasets do not contain specific fields or labels explicitly indicating `ambiguity' in queries.  As for metrics, similar to ACQ, BLEU and ROUGE are popular choices for measuring the quality of rewritten queries. Additionally, since QR is often employed for IR tasks, standard IR metrics such as mean reciprocal rank (MRR), mean average precision (MAP), and Recall@k and Precision@k are used to evaluate whether the rewritten query retrieves the correct information \cite{ma2023query, ye2023enhancing}. For these purposes, popular open-domain QA datasets like NQ \cite{kwiatkowski2019natural}, TriviaQA \cite{joshi-etal-2017-triviaqa}, and HotpotQA \cite{yang2018hotpotqa} are often used as benchmarks. However, we do not list them here as they do not focus specifically on ambiguity and lack corresponding human-rewritten queries.

\begin{table*}[h!]  
\centering  
\resizebox{\textwidth}{!}{
\begin{tabular}{lccccc}  
\hline  
\textbf{Disambiguation Policy} & \textbf{Automatic?} & \textbf{Additional LLM Call?} & \textbf{Visible to User?} & \textbf{High Risk?} & \textbf{UX Disrupting?} \\ \hline  

Query Rewriting & \cellcolor{green!20}Yes & \cellcolor{red!20}Yes & \cellcolor{green!20}No & \cellcolor{red!20}Yes & \cellcolor{green!20}No \\

Long Form Answer Generation & \cellcolor{green!20}Yes & \cellcolor{yellow!20}Maybe & \cellcolor{red!20}Yes & \cellcolor{green!20}No & \cellcolor{yellow!20}Maybe \\

Asking Clarification Question & \cellcolor{red!20}No & \cellcolor{red!20}Yes & \cellcolor{red!20}Yes & \cellcolor{green!20}No & \cellcolor{red!20}Yes \\

\hline  
\end{tabular}
}
\caption{Comparison of disambiguation policies across key dimensions. Trait colors: \textcolor{green!80}{Green} = positive, \textcolor{red!60}{Red} = negative, \textcolor{yellow!50!orange}{Yellow} = context-dependent. No single policy suffices, motivating an agentic framework to coordinate.}
\label{tab:comparison}
\end{table*}

\noindent{\textit{\textbf{Long Form Answer Generation.}}} To the best of our knowledge, ASQA \cite{stelmakh-etal-2022-asqa} is the only dataset that falls into this category. ASQA is a long-form QA dataset derived from a subset of ambiguous questions in the AmbigNQ dataset \cite{min2020ambigqa}. Its statistics and corresponding URL are provided in \Cref{tab:ambig-datasets-cleaned}. The dataset is designed to evaluate how well systems can generate comprehensive answers that cover all valid interpretations. Two main metrics are used to assess the generation quality \cite{in2024diversify}: Disambig-F1 (D-F1) \cite{stelmakh-etal-2022-asqa}, which assesses the accuracy of responses by verifying correct answers to disambiguated questions using an F1 score, and ROUGE, which evaluates the correctness by comparing them to ground-truth long-form answers.

\section{Open Problems and Challenges}\label{sec:open-problems-challenges}
\subsection{Detecting Ambiguities}
While LLMs have exceptional generative capabilities, recent studies consistently highlight the challenges of using LLMs to detect ambiguous queries with high performance. For example, \citet{zhang2023clarify} achieved an AUROC of $0.57$ on AmbigNQ \cite{min2020ambigqa} using \texttt{LLaMA-2-13B-Chat}, while \citet{zhang2024clamber} reported a best F1 score of $0.53$ on their dataset using \texttt{GPT-3.5-Turbo}. \citet{tanjim2025detecting} shares a similar study and highlight a relatively lower performance using \texttt{GPT-3.5-Turbo} and \texttt{LLaMA-3.1-70B}. One potential reason, as suggested by \citet{liu2023we}, is that LLMs are not inherently designed to model ambiguities. 

\subsection{How To Orchestrate?}
This is one of the research questions we posed at the beginning. To first see why we need to ochestrate among the disambiguation policies, in this paper, we systematically analyze the pros and cons of each disambiguation policy, making us the first to do so to the best of our knowledge. We show the list in \Cref{tab:comparison}, which are: 
\textit{1) Automatic:} Both QR and LFAG are automatic and do not require human validation, unlike clarifying questions. \textit{2) Additonal LLM Call:} For CQA, at least one LLM call is needed for answer generation, and so LFAG could be integrated into that same LLM call. But both QR and ACQ require dedicated LLMs. \textit{3) Visible to User:} Rewritten queries are not typically visible to the user, whereas users might notice long-form answers and are definitely aware of clarifying questions. \textit{4) High Risk:} Each policy affects different processing steps; for example, QR impacts downstream tasks significantly, as incorrect assumptions can lead to wrong answers. \textit{5) UX Disrupting:} Repeated QR does not affect user experience as it is not visible, but too many clarifying questions can vex users. LFAG falls in between, as overly long answers are sometimes unwelcome. As can be seen, each approach has unique strengths and weaknesses, necessitating the need of coordination. The challenge lies in determining when to use which policy. For example, always asking clarifying questions can disrupt UX while always rewriting queries can lead to errors \cite{tanjim2025detecting}. These lead to multiple opportunities which we lay out below.

\subsection{Opportunities}
The next wave of disambiguation in Conversational QA is being shaped by three emerging trends: agentic orchestration,  simulation- and reward-driven policy optimization, and evaluation with LLM-as-a-judge. 
\vspace{0.2cm}

\noindent{\textit{\textbf{Agentic orchestration for disambiguation.}}}
Recent generations of LLMs \citep{meta2024llama32,openai2025o3,openai2025gpt5} offer longer context windows, more reliable tool use, and reasoning-centric architectures. These advances make it increasingly feasible to deploy \emph{multi-agent} CQA systems \citep{dibia2024autogen, fourney2024magentic}. Within such frameworks, disambiguation becomes a first-class capability for every agent, but especially for the orchestrator or coordinator agent, which explicitly handles user requests and is responsible for carrying forward the task with the help of other agents.

To operationalize this capability, the orchestrator can leverage agent cards or specifications in tandem with other agents to agree on common nomenclature or establish a constitution. By standardizing specification languages, these can systematically encode explicit policies for query handling and clarification, with \citet{agents_md} offering a promising step toward such standardization. For instance, an agent card may include a disambiguation extension that determines when to issue clarification prompts (e.g, which tools to use). Moreover, agentic orchestration can incorporate auxiliary mechanisms, such as memory and verification, to safeguard trajectory alignment. Memory management preserves relevant conversational history \citep{anthropic2025memory} while filtering out noise, thereby reducing the risk of context drift. LLM verification (discussed separately later) serves as an additional checkpoint, validating whether the system’s chosen path aligns with the intended query resolution.

Conversational engines can further enhance this process by engaging users in real-time clarification. Inspired by Bayesian Experimental Design (BED) \citep{rainforth2024modern}, a promising direction in this space, as explored in \citet{kobalczykactive2025}, is to actively select questions that maximize expected information gain, shifting from implicit reasoning about the best question to explicit evaluation via sampling from the solution space. Agents with such advanced meta-cognitive skills could eventually infer the most informative questions autonomously, combining static policies from agent cards with dynamic dialogue to robustly handle ambiguous input. Thus, embedding these strategies directly into the
orchestrator's as well as other agents' layer ensures that disambiguation is
not an afterthought, but a modular and transparent component of multi-agent coordination.
\vspace{0.2cm}

\noindent{\textit{\textbf{{Simulation- and reward-driven policy selection}}}}.
A major opportunity lies in training multiple disambiguation policies shown in \Cref{tab:comparison} using simulation. By generating large-scale ambiguous dialogues and optimizing reward-driven objectives, policies can be tuned not only for task accuracy but also for groundedness, efficiency, and user experience. Inspired by advances in reinforcement learning for reasoning using simulaiton \citep{guo2025deepseek}, such systems can test alternative strategies (clarify vs.\ rewrite vs.\ direct answer) and optimize routing controllers accordingly. 
\citet{mukherjee2025learningclarifyreinforcementlearning} push a more practical path forward: rather than relying on SFT or preference-based tuning — both burdened with extra hyper-parameters and indirect reward alignment — their approach shows that QA agents can cut straight to the goal with reward-weighted supervised fine-tuning. This new offline RL objective offers a practical step for fine-tuning an orchestrator to select disambiguation policies more effectively than specification or search-based approach outlined above.
Multi-agent RL further enables coordination between other agents with multiple roles (planner, retriever, checker) via shared objectives \citep{chen2025mmoarag}. 
\vspace{0.2cm}

\noindent{\textit{\textbf{{LLM-as-a-judge for user-centric evaluation.}}}}
Conventional evaluation metrics such as BLEU, ROUGE, and METEOR are not aligned with the goals of disambiguation, where success requires semantic correctness, underspecification resolution, and conversational coherence. Recent work demonstrates that LLM-as-a-judge can evaluate responses more holistically according to faithfulness, clarity, relevance, and conversational satisfaction \citep{zheng2023mtbench,gu2024llmasjudge, lee2025agentic}. Specialized judges can enable multilingual and domain-targeted assessment \citep{kim2023prometheus,pombal2025mprometheus}. Embedding such judges directly ``in the loop'' (i.e., LLM-verification mentioned earlier) can provide dense, rubric-based feedback not only for outputs but also for \emph{policies}---e.g., ``should a clarification have been asked here?'' This can align evaluation goals with end-user satisfaction and accelerate policy refinement. Beyond scoring final answers using traditional LLM-as-a-judge, \emph{agent-as-a-judge} is another promising direction, where evaluating agents can audit intermediate steps (e.g., query reformulations, retrieval choices, clarification turns) \citep{zhuge2024agentasjudge}. 
\vspace{0.2cm}

\noindent{\textit{\textbf{{Human factors and user experience.}}}}
User experience remains central to all disambiguation strategies. Adaptive clarification thresholds that are confidence- and risk-aware, \emph{persona-sensitive clarification styles}, and \emph{transparent attribution mechanisms} can directly improve user trust and satisfaction. Importantly, minimizing unnecessary interruptions while ensuring correctness is crucial. As LLMs continue to scale in reasoning, planning, and orchestration capabilities, we anticipate that agentic CQA systems will increasingly arbitrate between clarification, rewriting, retrieval, and direct answering and advance toward trustworthy, engaging, reliable \emph{multi-agent systems}.

\section{Conclusion}
In this paper, we have provided a comprehensive analysis of ambiguity and disambiguation in LLM-based CQA systems through three fundamental research questions. First, we have explored different types of ambiguity and proposed a unified taxonomy using three categories. We also highlighted the challenges of accurately detecting ambiguity, even with LLMs. Next, we have categorized various LLM-based disambiguation approaches and reviewed key benchmark datasets and metrics. Finally, we discussed open challenges and opportunities for LLM-based ambiguity detection and disambiguation strategies, particularly from agentic perspectives. By offering a comprehensive review of current research on ambiguities and disambiguation with LLMs, we hope our survey will contribute to the development of more robust and reliable LLM-based applications.

\section*{Limitations} In this work, we aimed to provide a comprehensive review and categorization of recent research on LLM-based ambiguity detection and disambiguation. Through our analysis, we identified three simplified categories of ambiguity types and three primary disambiguation techniques. However, this categorization is not exhaustive and may differ from other frameworks, which often use more granular or task-specific classifications. Despite our thorough literature review, it is possible that some recent or less-publicized works were overlooked, given the rapid advancements in this field. Additionally, our survey focused exclusively on ambiguity in Conversational Question Answering (CQA) tasks. In this survey, we did not cover other important NLP tasks, such as Natural Language Inference (NLI), Machine Translation (MT), Information Retrieval (IR), and Code Generation (e.g., NL2SQL), where ambiguities also arise and pose significant challenges. Future work could benefit from extending the scope to include these tasks, providing a more holistic understanding of ambiguity in NLP applications.

\section*{Acknowledgements}

This work was supported by Institute of Information \& Communications Technology Planning \& Evaluation(IITP) grant funded by the Korea government(MSIT) (RS-2023-00216011), as well as another IITP grant funded by the Korea government(MSIT) (RS-2022-II220077).

\bibliography{custom}
\end{document}